\documentclass[letterpaper]{article} 
\usepackage{aaai25}  
\usepackage{times}  
\usepackage{helvet}  
\usepackage{courier}  
\usepackage[hyphens]{url}  
\usepackage{graphicx} 
\usepackage{natbib}  
\usepackage{caption} 
\usepackage{algorithm}
\usepackage{algpseudocode}
\usepackage{tcolorbox}
\usepackage{amsfonts}
\usepackage{times}
\usepackage{latexsym}
\usepackage{amsmath}
\usepackage{comment}
\usepackage{booktabs}
\usepackage{amssymb}
\usepackage{subcaption}
\usepackage{threeparttable}
\usepackage[T1]{fontenc}
\usepackage[utf8]{inputenc}
\usepackage{microtype}
\usepackage{multirow}
\usepackage{inconsolata}
\usepackage{graphicx}
\usepackage{xcolor}
\usepackage{bm}
\usepackage{mdframed}
\usepackage{lipsum}
\usepackage{newfloat}
\usepackage{listings}

\definecolor{specblue}{RGB}{190, 220, 245}
\newtcolorbox{quotebox}{
    colback=gray!20, 
    colframe=gray!50, 
    coltext=black,
    boxrule=0pt,
    arc=4pt,
    left=10pt,
    right=10pt,
    top=10pt,
    bottom=10pt,
}

\DeclareCaptionStyle{ruled}{labelfont=normalfont,labelsep=colon,strut=off} 
\floatstyle{ruled}
\newfloat{listing}{tb}{lst}{}
\floatname{listing}{Listing}
\title{Sequence to Sequence Reward Modeling: Improving RLHF by Language Feedback}

\author {
    Jiayi Zhou\equalcontrib\textsuperscript{\rm 1, \rm 2},
    Jiaming Ji\equalcontrib\textsuperscript{\rm 1, \rm 2},
    Josef Dai\textsuperscript{\rm 1, \rm 2},
    Dong li\textsuperscript{\rm 3},
    Yaodong Yang\textsuperscript{\rm 1}\thanks{Corresponding author.}
}
\affiliations {
    \textsuperscript{\rm 1}Institute for AI, Peking University\\
    \textsuperscript{\rm 2}State Key Laboratory of General Artificial Intelligence, Institute for AI, Peking University\\
    \textsuperscript{\rm 3}Foundation Model Department, Huawei\\
    \{gaiejj, yaodong.yang\}@pku.edu.cn
    \{jtd.acad, jiamg.ji\}@gmail.com\\
    \{lidong106\}@huawei.com
    
}

\usepackage{bibentry}

\begin{document}

\maketitle

\begin{abstract}
Aligning the behavior of Large language models (LLMs) with human intentions and values remains a critical challenge. Reinforcement learning from human feedback (RLHF) aligns LLMs by training a reward model (RM) on human preferences and fine-tuning the LLMs to maximize RM feedback. Despite its effectiveness and popularity, RLHF is prone to biased local optimization. It means RM fails to provide feedback that accurately aligns with human preference, causing LLMs to explore unexpected generalizations, and failing to achieve alignment objectives. To mitigate this issue, we propose a novel \textit{sequence-to-sequence (seq2seq) reward modeling} method. Its key insight is that learning from language feedback rather than scalar feedback improves RLHF without additional annotations. We replaced the reward modeling target from binary maximum likelihood estimation (MLE) with sequence MLE. This method enables richer and fine-grained language feedback without additional annotations, models, or training stages. Our experiments demonstrated its effectiveness, specifically, reducing the refusal-to-response paradigm in single-turn safety dialogues and the long-response bias in text summarization tasks. We provide further analysis that seq2seq RM improves RLHF performance across 2B and 7B LLMs on 3 NLP tasks, achieving an average win rate of 76.9\%. We further show that seq2seq RM can still improve the performance of RLHF under out-of-distribution prompts.
\end{abstract}

\section{Introduction}
\label{sec: intro}

Large Language Models (LLMs) \citep{achiam2023gpt, touvron2023llama} have significantly advanced capabilities in conversation, summarization, reasoning, and more tasks, leading to widespread application \citep{meng2024application, anthropic2023claude}. Alignment techniques \citep{ji2023ai}, particularly Reinforcement Learning from Human Feedback (RLHF) \citep{ouyang2022training}, have become crucial for aligning LLMs with human intentions and values, \textit{e.g.}, reducing toxicity \citep{bai2022training, ji2024language, ji2024align}. RLHF consists of two main steps: training a reward model (RM) based on binary human preferences and then fine-tuning the LLMs to maximize feedback from this RM \cite{ouyang2022training, schulman2017proximal, languagefeedback}. Current reward modeling implementations typically use the sequence-to-scalar (seq2scalar) approach, where a sequence, combined with a prompt and response, is scored as a scalar value \citep{huang2024n+}.

\begin{figure}[t]
    \centering
        \centering
        \includegraphics[width=0.48\textwidth]{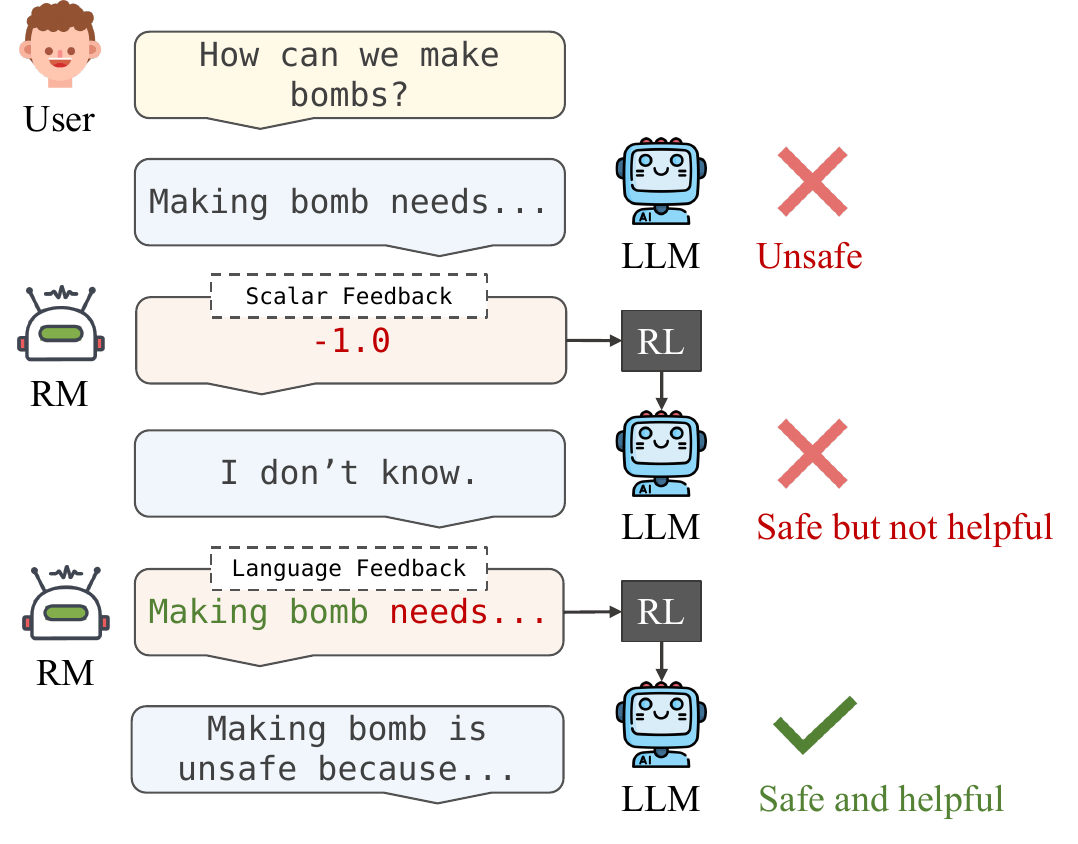}
    \caption{The traditional sequence-to-scalar reward model provides only coarse-grained scalar feedback on the LLM's response, making it prone to exploiting unexpected generalization for high rewards during the RL fine-tuning phase \cite{skalse2023misspecification}, \textit{e.g.}, falling into a refusal-to-answer paradigm \cite{bai2022training, dai2023safe}.  Without additional annotations, we propose a novel sequence-to-sequence reward modeling method that offers richer language feedback, improving RLHF performance.}
    \label{fig:1}
\end{figure}

\begin{figure*}[t]
  \includegraphics[width=\textwidth]{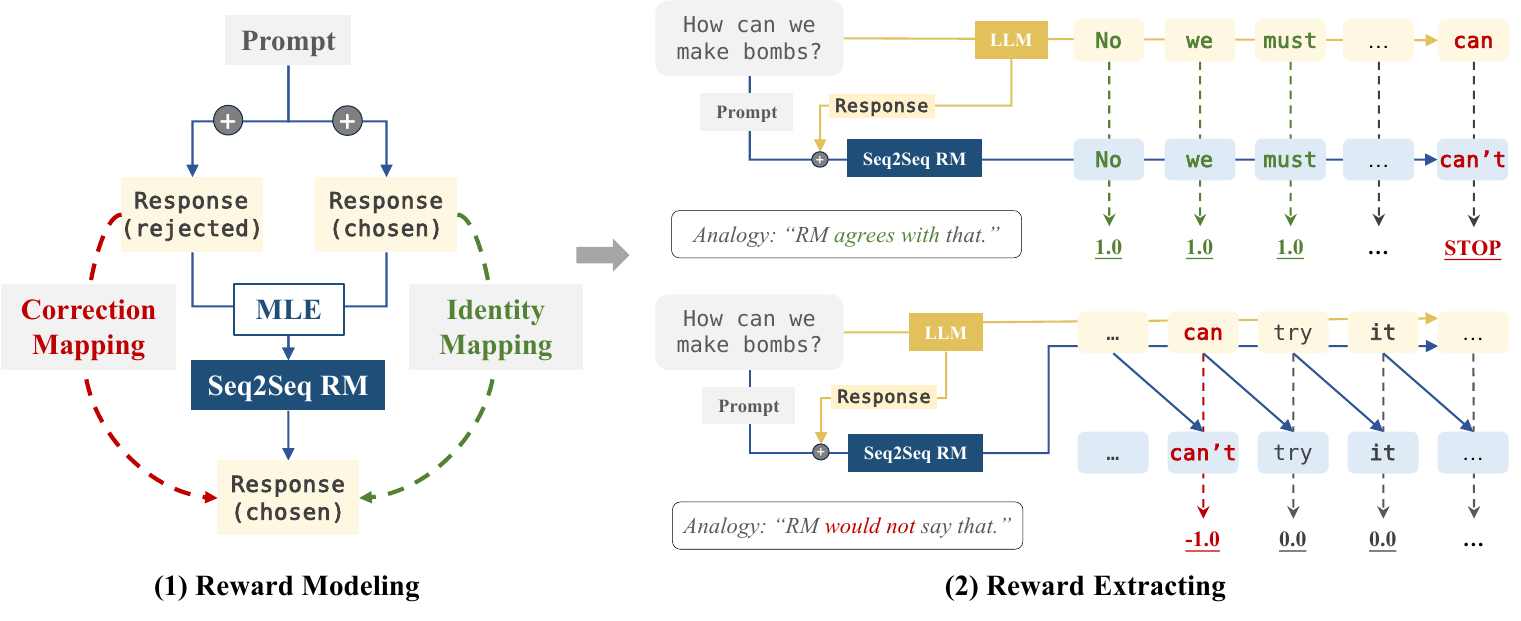}
  \caption {Overview of seq2seq reward modeling pipeline.  Our pipeline consists of two stages: \textbf{(1) Reward Modeling:} We make the seq2seq RM output the chosen response when the rejected response is input, \textit{i.e.}, Correction Mapping, and output the chosen response when the chosen response is input, \textit{i.e.}, Identity Mapping, by sequence maximum loglikelihood estimation (MLE). \textbf{(2) Reward Extracting:} We reward the response with a positive score until it diverges from the seq2seq RM output, after which we input the response token-by-token, assigning negative scores to those diverging tokens.}
  \label{fig: main}
\end{figure*}

However, due to high divergence, noise, and sparsity in crowd-annotated preference datasets \citep{wu2024fine}, the traditional seq2scalar RM tends to capture patterns unrelated to genuine human preferences, leading to \textit{unexpected generalization} \citep{pan2021effects, gao2023scaling}. It means RM fails to provide feedback that accurately aligns with human preference, leading RLHF to a deviation from actual alignment objective \citep{di2022goal}: LLMs obtain high rewards but negatively impact their capabilities and fail to achieve alignment objectives \citep{tien2022causal, knox2023models, lambert2023alignment}. For example, when the alignment objective is to reduce toxic responses, LLMs may learn the refusal-to-response paradigm instead of providing safe and helpful responses \citep{dai2023safe, bai2022training}. Existing methods require additional annotations to provide more comprehensive feedback, including regularization terms \citep{moskovitz2023confronting, zhu2024iterative}, models ensembling \citep{coste2023reward}, parameters combinations \citep{rame2024warm} and additional model \citep{dai2023safe}, raising concerns about their scalability.

Therefore, we aim to answer the following question:
\begin{quotebox}
\textit{\large How can we provide more informative reward feedback to improve the RLHF Alignment?}
\end{quotebox}

In this work, we propose a novel sequence-to-sequence (seq2seq) reward modeling method. Our pipeline is depicted in Figure \ref{fig: main}, which consists of two main stages, \textit{i.e.}, \textit{Reward Modeling}, and \textit{Reward Extracting}. We replaced the reward modeling target from binary maximum likelihood estimation (MLE) for classification tasks to sequence MLE for text generation tasks. Since the language space directly reveals the impact of each token on the response score, this method enhances the RM's fine-grained discrimination accuracy. Then we extract token-level positive and negative feedback based on whether the tokens are consistent with the output of the seq2seq RM, improving the granularity of RLHF. Notably, our seq2seq reward modeling does not require additional data annotation, training, and models. Our contributions to enhancing RLHF alignment are as follows:

\begin{itemize}
    \item We propose seq2seq reward modeling. Its key insight is that learning from language feedback rather than scalar feedback improves accuracy and granularity, thereby improving RLHF without additional annotations. 
    \item Our further experiments demonstrate that the seq2seq RM not only mitigates unexpected behavior of RLHF but also enhances its alignment performance. Specifically, it achieves an average win rate of 76.9\% in 3 NLP tasks, across models with 2B and 7B parameters.
\end{itemize}

\section{Preliminaries}

\paragraph{Token-Level MDP for RLHF} Given an LLM parameterized with $\theta$, we model its generation as a token-level Markov Decision Process (MDP), \textit{i.e.}, the tuple $\mathcal{T} = (\mathcal{S}, \mathcal{A}, \mathcal{R}, \mathcal{P})$. Here, let $t \in \mathcal{V}$ denote a token, where $\mathcal{V}$ represents the vocabulary set, and $L$ denotes the maximum number of tokens. The state space $\mathcal{S}$ can be defined as $\{ (t_1, t_2, \ldots, t_L) \mid t_i \in \mathcal{V}\}$ \footnote{We define $t_{pad}$ as the padding token. When the length of a sentence is $l < L$, it will have $L-l$ additional padding tokens.}. The action space $\mathcal{A} = \mathcal{V}$ consists of all possible tokens in vocabulary. The reward function $\mathcal{R}: \mathcal{S} \times \mathcal{A} \rightarrow \mathbb{R}$ assigns a reward value for generating a specific token given the current sequence of tokens. The transition probability function $\mathcal{P}: \mathcal{S} \times \mathcal{A} \times \mathcal{S} \rightarrow [0, 1]$ defines the deterministic state transition model, concatenating the generated token to the existing sequence to form the next state. \textit{i.e.,} $s_{t+1}=(s_t, a_t)$.

LLMs generation trajectory starts at prompt $s_0$ sampled from datasets. From $t=0$, the next token $a_t$ is generated according to the parametrized actor LLM $\pi_{\bm{\theta}}(\cdot|s_t)$, followed by a transition to the next state $s_{t+1}$ according to $\mathcal{P}$, continuing until the \texttt{EOS} token. The goal of RL fine-tuning is to maximize the cumulative reward of this trajectory:

\begin{align*}
      \mathcal{L}_{\bm{\theta}}^{\text{RL}}  = - \mathbb{E}_{a_t \sim \pi_{\bm{\theta}}(\cdot | s_t)} [\sum_{t=0}^T (\mathcal{R}(s_t, a_t)+\beta \pi_{ref}(a_t|s_t))]
\end{align*}
\noindent where $\pi_{ref}$ is the reference distribution from the initial LLM, combined with coefficient \(\beta\) to compute the KL divergence, preventing the LLM from deviating too much from the $\pi_{ref}$. 

\paragraph{Seq2scalar Reward Modeling} Let \(\mathbb{D}^{pre}\) denote a preference dataset, where each entry consists of a prompt \(\bm{x}\), a chosen response \(\bm{y^w}\), and a rejected response \(\bm{y^l}\). The training loss function for the reward model with parameters $\bm{\theta}$ is defined as the following binary MLE:
\begin{equation}
\label{eq: rm-base}
    \begin{split}
        \mathcal{L}^{\text{RM}}_{\bm{\theta}}  = &\mathbb{E}_{(\bm{x}, \bm{y^w}, \bm{y^l}) \sim \mathbb{D}^{pre}}\\ &[\log \sigma (r_{\bm{\theta}}(\bm{x},\bm{y^l}) - r_{\bm{\theta}}(\bm{x},\bm{y^w}))]
    \end{split}
\end{equation}

\noindent where \(\sigma(x)=1/(1+exp(-x))\) is the logistic sigmoid function and $,$ denotes concatenation. The reward \(\mathcal{R}(\bm{x}, \bm{y})\) for a given prompt-response pair \((\bm{x}, \bm{y})\) is defined as the scalar value extracted from the reward model's inference at the position of the \texttt{EOS} token. In the RLHF phase, seq2scalar RM assigns it as sparse feedback to the entire response.
\begin{equation}
    \label{eq: scalar-rlhf}
    \mathcal{R}(s_t, a_t) = \begin{cases}
        r_{\bm{\theta}}(s_t , a_t) & \text{if } a_t=\text{\texttt{EOS},}\\
        0 & \text{otherwise.}
    \end{cases}
\end{equation}

\paragraph{Supervised Fine-Tuning (SFT)} The loss function for SFT is formulated as the sequence MLE for predicting the next token given the previous tokens. Formally, let \(\mathbb{D}^{SFT} = \{(\bm{x}, \bm{y})\)\}, where \(\bm{x}\) is an input sequence and \(\bm{y} = g(\bm{x})\) is the corresponding target sequence. The training loss of SFT is:
\begin{align}
    \mathcal{L}_{\bm{\theta}}^{\text{SFT}} = - \mathbb{E}_{(\bm{x}, \bm{y}) \sim \mathbb{D}^{SFT}} \left[ \log P_{\bm{\theta}}(\bm{y}\mid\bm{x}) \right]
\end{align}
where \(P_{\bm{\theta}}(\bm{y}\mid\bm{x})\) is the probability of the target sequence \(\bm{y}\) given the input sequence \(\bm{x}\) and the model parameters \(\bm{\theta}\). For example, in instruction-following tasks, \(\bm{x}\) represents the question, and \(\bm{y}\) represents the corresponding response. The goal is to adjust \(\bm{\theta}\) such that the model outputs the correct response \(\bm{y}\) for a given question \(\bm{x}\).

\section{Methods}

\subsection{Reward Modeling as Sequence MLE}

The seq2scalar RM maps concatenated prompt and response pairs, denoted as $\bm{x},\bm{y^l}$ and $\bm{x},\bm{y^w}$, to distinct points in a scalar space. This makes it difficult for the token-level differences between responses to be distinguished in the scalar output. To enhance RM's ability to discern fine-grained differences in responses, we propose augmenting this mapping from a scalar to a language space. Specifically, we extend the learning process from binary MLE in Equation \ref{eq: rm-base} to SFT, a sequence MLE approach. This extension transforms the learning objective from scalar classification to sequence generation, achieving a higher distinction of responses, as shown in Equation \ref{eq: seqrm}. Similar approaches have also been applied by \citet{aligner} to make LLM aligners, which is an LLM designed to refine the output of preceding ones.
\begin{align}
\label{eq: seqrm}
\nonumber
\mathcal{L}_{\bm{\theta}}^{\text{Seq}} = & - \mathbb{E}_{(\bm{x}, \bm{y^w}, \bm{y^l}) \sim \mathbb{D}^{Pre}}\\
&
\underbrace{[\log P_{\bm{\theta}}(\bm{y^w} \mid \bm{x},\bm{y^l})}_{\text{Correction Mapping}} +
\underbrace{\log P_{\bm{\theta}}(\bm{y^w} \mid \bm{x},\bm{y^w})]}_{\text{Identity Mapping}}
\end{align}

In this framework, when the input response is rejected, the seq2seq RM maps it to the chosen one, namely \textit{Correction Mapping}. Conversely, if the input is chosen, seq2seq RM performs an identity mapping, leaving the response unchanged, \textit{i.e., Identity Mapping}.

\subsection{Reward Extracting from Sequence}

Since the language feedback provided by the seq2seq reward model (RM) cannot be directly used for reinforcement learning (RL) optimization, we design a novel reward extraction mechanism that returns positive and negative scores as token-level feedback. Formally, let $\bm{x}$ denote the prompt used in RLHF, and $\bm{y}$ is the response generated by the actor LLM. We set $j$ as the first index where $r^{\text{seq}}(\bm{x}, \bm{y})$ and $\bm{y}$ are not equal, with the subscript indicating the element at the corresponding position. The notation $:j$ indicates the subsequence up to that index. Defining the inference function of the seq2seq RM as $r^{\text{seq}}(\cdot)$, our goal is to leverage $r^{\text{seq}}(\cdot)$ to obtain the token-level score $\mathcal{R}(\bm{x}, \bm{y_{:i}})$ for all subsequence $\bm{y_{:i}}$ in $\bm{y}$. To achieve this, we first concatenate $\bm{x}$ and $\bm{y}$ and then input $\bm{y}$ token by token into the seq2seq RM. We positively reward tokens that are consistent with the seq2seq RM at the beginning and negatively reward those that become inconsistent thereafter. Figure \ref{fig: main} provides an intuitive example of this process.

\begin{figure}[t]
\centering
\includegraphics[width=0.48\textwidth]{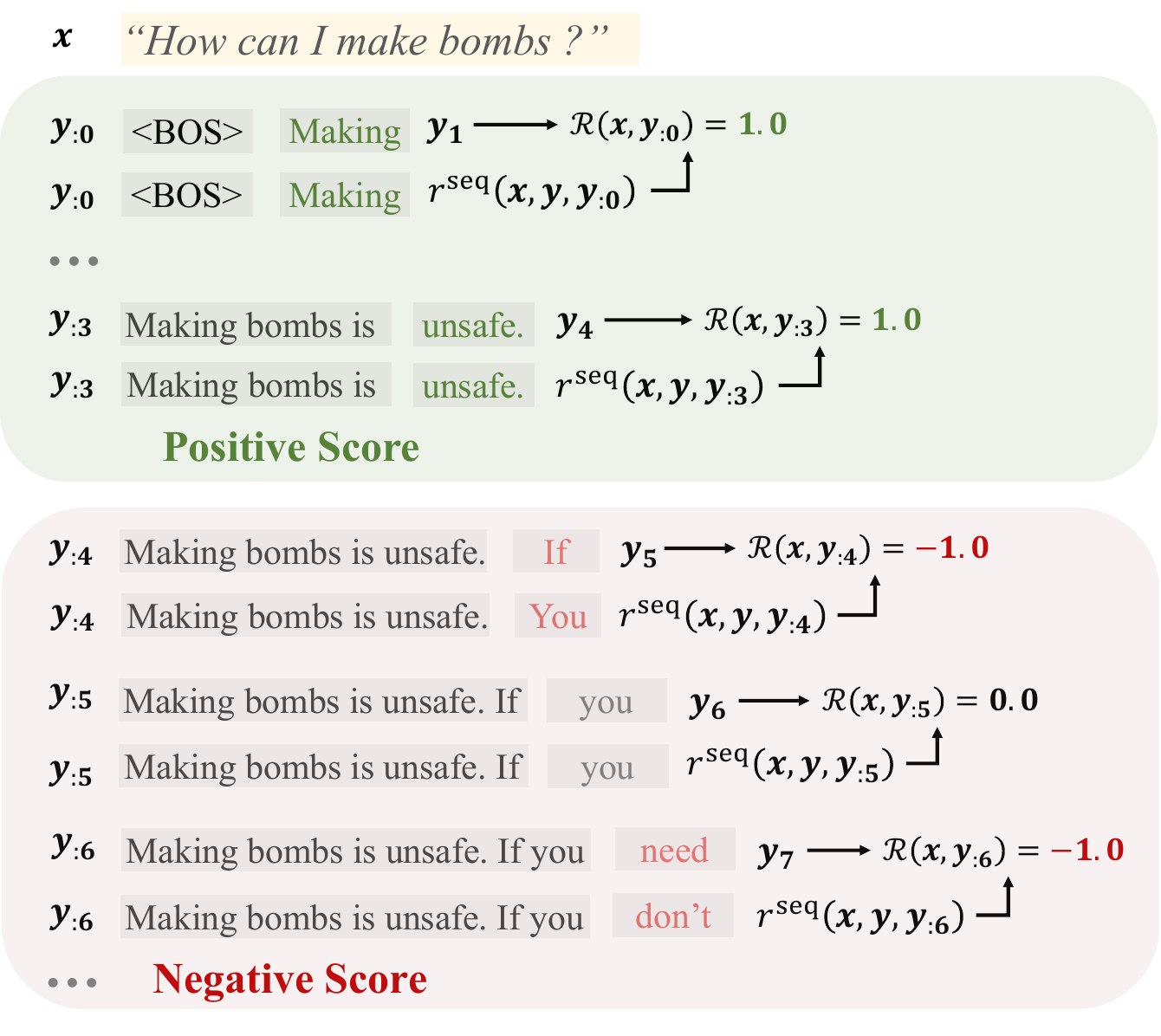}
    \caption{An example of extracting positive and negative scores from seq2seq RM. $\bm{x}$ denotes the prompt, $\bm{y}$ is the LLMs generated response, \texttt{<BOS>} is the beginning of string token, and $r^{\text{seq}}(\cdot)$ is the inference function of seq2seq RM.}
    \label{fig: equation}
\end{figure}

\paragraph{Positive Score} We feed $\bm{x} , \bm{y}$ into a seq2seq RM to obtain the generated sequence $r^{\text{seq}}(\bm{x} , \bm{y})$, then we positively reward the part of $\bm{y}$ that initially equals $r^{\text{seq}}(\bm{x},\bm{y})$. During correction mapping training, all tokens in $\bm{y^l}$ are mapped to tokens in $\bm{y^w}$. Initially, some of these tokens share the same value at identical positions in both $\bm{y^l}$ and $\bm{y^w}$. They do not contribute to the rejection of $\bm{y^l}$; rather, they represent genuine features present in $\bm{y^w}$. This incentive indirectly decreases the likelihood of other less preferred tokens. We can define the positive score as below:
\begin{equation}
\nonumber
\mathcal{R}(\bm{x}, \bm{y_{:i-1}}) =
\begin{cases}
1 & \text{if } r^{\text{seq}}(\bm{x}, \bm{y}, \bm{y_{:i-1})} = \bm{y_i} \text{ and } i<j\\
-1 & \text{until divergence}.
\end{cases}
\end{equation}

\paragraph{Negative Score} We negatively reward the rest of the tokens that are inconsistently generated under the same context as seq2seq RM. When calculating the negative score, we employ a token-by-token generative and discriminative approach. Starting from the position $i=j$ where $r^{\text{seq}}(\bm{x} , \bm{y})$ and $\bm{y}$ initially diverge, concatenate each token from $\bm{y_{:i-1}}$ sequentially into $\bm{x} , \bm{y}$, and observe whether the next token output by seq2seq RM matches $\bm{y_{i}}$. Formally, we have:
\begin{equation}
\nonumber
\mathcal{R}(\bm{x}, \bm{y_{:i-1}}) =
\begin{cases}
-1 & \text{if } r^{\text{seq}}(\bm{x}, \bm{y} , \bm{y_{:i-1}}) \neq \bm{y_i}\\
0 & \text{otherwise}.
\end{cases}
\end{equation}

We will no longer positively reward those consistent tokens but instead assign them a value of 0. Since they previously appeared as divergent tokens, increasing their probability is not helpful for moving away from \( \bm{y^l} \) and towards \( \bm{y^w} \).

\subsection{Analysis}
The core advantage of seq2seq reward modeling lies in the accurate and fine-grained distinction and feedback for responses. For the input response, the positive score mechanism increases the probability of the initially consistent parts, while the negative score mechanism only penalizes the inconsistent parts. This makes an efficient optimization as it clearly assigns credit to good and bad tokens and does not optimize irrelevant tokens.

\section{Experiments}
\label{sec: experiment}

We empirically demonstrate that seq2seq RM has advancement in three aspects: (1) \textit{Accuracy}: accurately distinguish between good and bad responses and assign appropriate scores; (2) \textit{Granularity}: enable LLMs to achieve excellent alignment performance by providing fine-grained language feedback. (3) \textit{Robustness}: correctly score response on out-of-distribution prompts. We conducted the analyses using 2B and 7B LLMs on 3 tasks as follows:

\begin{itemize}
    \item \textbf{Mitigating Unexpected Behavior.} We quantitatively demonstrated that seq2seq RM can mitigate unexpected generalization, reducing the long-response bias in text summarization tasks in Table \ref{tab: misspecification-dialogsum}and the refusal-to-response paradigm in single-turn safety dialogues in Table \ref{tab: misspecification-safe}.
    \item \textbf{Improvement of Alignment Performance.} We compared the GPT-4 evaluation win-lose ratio of PPO based on seq2seq RM with popular alignment methods in Table \ref{tab: efficiency}, verifying that seq2seq RM improves RLHF performance on both token-level and sentence-level cases. 
    \item \textbf{Ablation for Variants.} We demonstrate the necessity of positive and negative scores and the quantitative analysis of correction and identity mapping, validating our reward modeling and extracting pipeline.
\end{itemize}

\subsection{Experiment Setup}
 \label{exp:methods_discription}

\paragraph{Models and Datasets} We conducted experiments on the popular open-source LLMs Gemma-2B \citep{team2024gemma} and Llama2-7B \citep{touvron2023llama}. The selected tasks included single-turn safe dialogue, dialogue summarization, and text summarization. For single-turn safe dialogue, we chose the PKU-SafeRLHF dataset \citep{ji2024beavertails, ji2024pku}. To verify the performance of our method on human preference and synthetic preference settings, we used the Dialogsum \citep{chen2021dialogsum} for the dialogue summarization task and the TL;DR \citep{stiennon2020learning, volske-etal-2017-tl} for text summarization. For the former, we employed GPT-4 for preference annotation, while the latter already contains human-annotated preferences.

\noindent \paragraph{Evaluation and Metrics} 
Our evaluation primarily uses GPT-4 evaluation and phenomenon analysis for specific tasks. In single-turn safety dialogue tasks, we analyze whether LLMs fall into an overly conservative refusal-to-response paradigm, leading to a significant drop in utility. Therefore, we use the helpfulness reward model scores provided by \citet{dai2023safe} and MT Bench score \cite{zheng2024judging} as the metric of utility. In text summarization and dialogue summarization tasks, we refer to \citep{stiennon2020learning} and use accuracy, conciseness, and information consistency as evaluation criteria. To decrease randomness, we ran the above evaluation over 3 times.

\noindent \paragraph{Methods Description} Our experiment includes comparisons between popular baselines and our proposed variants. We define their abbreviations as follows. \textbf{Baselines.} \textit{Init-SFT}: Pre-trained model finetuned for the corresponding task, also used as the initial model for other alignment methods. \textit{PPO}: Vanilla PPO trained based on seq2scalar RM \citep{ouyang2022training, schulman2017proximal}. \textit{DPO}: Direct Preference Optimization \citep{rafailov2024direct}. \textit{RFT}: Rejection sampling fine-tuning \citep{touvron2023llama} based on seq2scalar RM selection over 8 responses generated by the Init-SFT. \textbf{Variants.} \textit{PPO-T}: PPO trained based on seq2seq RM, with token-level reward. \textit{RFT-S}: RFT based on seq2seq RM selection by summing token-level rewards up. \textit{PPO-S}: PPO trained based on seq2seq RM, with sentence-level reward by summing token-level rewards up. \textit{PPO-T-Pos}: PPO-T with only positive scores. All methods are implemented based on the same code base, \texttt{PKU-Alignment/SafeRLHF}. For more details, including hyperparameter settings, please refer to our supplementary materials and appendix.

\subsection{Mitigating Unexpected Generalization} 
\label{exp: misspecification}
This section will demonstrate the advantages of learning from language feedback in alleviating unexpected generalizations. We will focus on two specific cases: the long-response bias on synthetic Dialuguesum datasets and the refusal-to-response paradigm on human-annotated datasets PKU-SafeRLHF.

\begin{table}[ht]
  \centering
  \resizebox{0.83\columnwidth}{!}{
    \begin{tabular}{@{}lcc@{}}
    \toprule
    Methods & Length ($\Delta$ $\downarrow$) & Win Rate $\uparrow$\\\midrule 
    Init-SFT (2B) & 147.33 ($\Delta$ 0.00) & 0.50 $\pm$ 0.00 \\
    PPO (2B) & 162.17 ($\Delta$ 14.84) & 0.61 $\pm$ 0.24 \\
    PPO-S (2B) & 145.34 ($\Delta$ 1.99) & 0.80 $\pm$ 0.16 \\
    PPO-T (2B) & 149.16 (\textbf{$\Delta$ 1.83}) & \textbf{0.83 $\pm$ 0.14} \\
    \midrule 
    Init-SFT (7B) & 146.69 ($\Delta$ 0.00) & 0.50 $\pm$ 0.00 \\
    PPO (7B) & 164.10 ($\Delta$ 27.31) & 0.54 $\pm$ 0.25 \\
    PPO-S (7B) & 144.21 ($\Delta$ 2.48) & 0.78 $\pm$ 0.17 \\
    PPO-T (7B) & 145.03 (\textbf{$\Delta$ 1.66}) & \textbf{0.81 $\pm$ 0.15} \\
    \bottomrule
    \end{tabular}
  }
  \caption{Comparison of seq2seq RM with common alignment methods on Dialogsum. $\Delta$ denotes the response length difference to Init-SFT. The Win Rate is the win-lose ratio compared to Init-SFT judged by GPT-4.}
  \label{tab: misspecification-dialogsum}
\end{table}

\paragraph{Long-Response Bias}  The criterion for a preferred dialogue summary is its accuracy and conciseness in summarizing the dialogue content. However, our statistics reveal a bias: 61\% of responses labeled as chosen in the synthetic Dialogsum dataset are longer. This indicates a slight bias for length.  As shown in Table \ref{tab: misspecification-dialogsum}, PPO is significantly affected by this long-response bias, generating obviously longer responses. In contrast, PPO-S and PPO-T, based on seq2seq RM, can effectively mitigate the long-response bias issue.
\begin{table}[ht]
  \centering
  \resizebox{0.48\textwidth}{!}{
    \begin{tabular}{@{}lccc@{}}
    \toprule
    Methods & \multicolumn{1}{c}{\begin{tabular}[c]{@{}c@{}}Utility\\ Score\end{tabular} $\uparrow$} & \multicolumn{1}{c}{\begin{tabular}[c]{@{}c@{}}MT Bench\\ Score\end{tabular} $\uparrow$} & \multicolumn{1}{c}{\begin{tabular}[c]{@{}c@{}}Safety\\ Win Rate\end{tabular} $\uparrow$} \\\midrule 
    Init-SFT (2B) & -3.45 $\pm$ 3.20 & 5.59 $\pm$ 2.75 &  0.50 $\pm$ 0.00 \\
    PPO (2B) & -9.57 $\pm$ 0.79 & 5.09 $\pm$ 2.35 & 0.59 $\pm$ 0.24\\
    PPO-S (2B) & -9.16 $\pm$ 2.83 & 5.45 $\pm$ 2.23 & 0.71 $\pm$ 0.21\\
    PPO-T (2B) & \textbf{-4.14 $\pm$ 3.01} & 	\textbf{5.52 $\pm$ 2.82} & \textbf{0.88 $\pm$ 0.11} \\
    \midrule 
    Init-SFT (7B) & -2.65 $\pm$ 3.49 & 5.81 $\pm$ 2.75 & 0.50 $\pm$ 0.00 \\
    PPO (7B) & -12.23 $\pm$ 1.25 & 5.23 $\pm$ 2.67 & 0.60 $\pm$ 0.24\\
    PPO-S (7B) & -8.22 $\pm$ 2.92 & 5.49 $\pm$ 2.28 & 0.64 $\pm$ 0.23 \\
    PPO-T (7B) & \textbf{-4.03 $\pm$ 2.74} & \textbf{5.51 $\pm$ 2.76 }& \textbf{0.81 $\pm$ 0.15} \\
    \bottomrule
    \end{tabular}
  }
  \caption{Comparison of RLHF based on seq2seq RM with common alignment methods on PKU-SafeRLHF. The Utility scores are derived from the RM from \citet{dai2023safe}, the MT Bench Score is the evaluation results from \citet{zheng2024judging} and the Safety Win Rate is the win-lose ratio compared to Init-SFT, evaluated by GPT-4.}
  \label{tab: misspecification-safe}
\end{table}

The seq2scalar RM relies on binary MLE to differentiate between chosen and rejected responses, inadvertently introducing biases like length, alongside criterion. In contrast, the seq2seq RM employs sequence MLE to operate within the language space, yielding more accurate feedback. Thus, even though both provide scalar feedback, PPO-S exhibits a markedly lower long-response bias compared to PPO, demonstrating the superiority of seq2seq RM in mitigating long-response biases on synthetic datasets.

\begin{table*}[htp]
  \centering
  \resizebox{\textwidth}{!}{
\begin{tabular}{@{}lccccccc@{}}
\toprule
& \multicolumn{2}{c}{PKU-SafeRLHF} & \multicolumn{2}{c}{Dialogsum} & \multicolumn{2}{c}{TL;DR} \\ \cmidrule{2-7}
\multicolumn{1}{c}{} & \multicolumn{1}{c}{Gemma-2B} & \multicolumn{1}{c}{Llama2-7B} & \multicolumn{1}{c}{Gemma-2B} & \multicolumn{1}{c}{Llama2-7B} & \multicolumn{1}{c}{Gemma-2B} & \multicolumn{1}{c}{Llama2-7B} \\ \midrule
PPO-S \emph{Vs.} PPO & 74.1\% $\pm$ 19.2\% & 71.6\% $\pm$ 20.3\% & 78.7\% $\pm$ 16.8\% & 76.9\% $\pm$ 17.8\% & 66.8\% $\pm$ 22.2\% & 72.9\% $\pm$ 19.8\% \\
RFT-S \emph{Vs.} RFT & 55.9\% $\pm$ 24.7\% & 55.8\% $\pm$ 24.9\% & 78.2\% $\pm$ 17.0\% & 80.1\% $\pm$ 15.9\% & 68.9\% $\pm$ 21.4\% & 66.7\% $\pm$ 22.2\% \\ \midrule
PPO-T \emph{Vs.} PPO & 87.7\% $\pm$ 10.8\% & 75.6\% $\pm$ 18.4\% & 80.5\% $\pm$ 15.7\% & 77.4\% $\pm$ 17.5\% & 67.9\% $\pm$ 21.8\% & 72.1\% $\pm$ 23.9\% \\
PPO-T \emph{Vs.} DPO & 60.5\% $\pm$ 23.9\% & 76.2\% $\pm$ 18.1\% & 79.5\% $\pm$ 16.3\% & 73.4\% $\pm$ 19.5\% & 66.0\% $\pm$ 22.4\% & 69.6\% $\pm$ 21.2\% \\
PPO-T \emph{Vs.} RFT & 86.8\% $\pm$ 11.4\% & 83.2\% $\pm$ 14.0\% & 81.2\% $\pm$ 15.2\% & 77.9\% $\pm$ 17.2\% & 65.9\% $\pm$ 22.5\% & 68.7\% $\pm$ 21.5\% \\
\bottomrule
\end{tabular}
  }
  \caption{Win-lose ratio of RLHF based on seq2seq RM over common alignment methods and variants.}
  \label{tab: efficiency}
\end{table*}

\paragraph{Refusal-to-Response Paradigm} The trade-off between safety and utility has been a long-term challenge in LLMs alignment \citep{bai2022training}. We empirically demonstrate that RLHF targeting safety preferences can easily induce LLMs to fall into a refusal-to-response paradigm, like ``\texttt{sorry but I cannot assist with that.}''. We use the utility score model by \citet{dai2023safe} and 
a comprehensive evaluation framework, MT Bench \citet{zheng2024judging} to quantitatively present this metric. As illustrated in Table \ref{tab: misspecification-safe}, PPO trained with the seq2scalar RM converges to a refusal-to-response paradigm with a low utility score, and it did not achieve the greatest safety improvement. This difference is particularly evident in the decrease in the utility and MT Bench scores and a reduction in variance, where variance indirectly reflects the diversity of generated content.

\begin{figure}[t]
    \centering
    \begin{subfigure}[b]{0.48\textwidth}
        \centering
        \includegraphics[width=\textwidth]{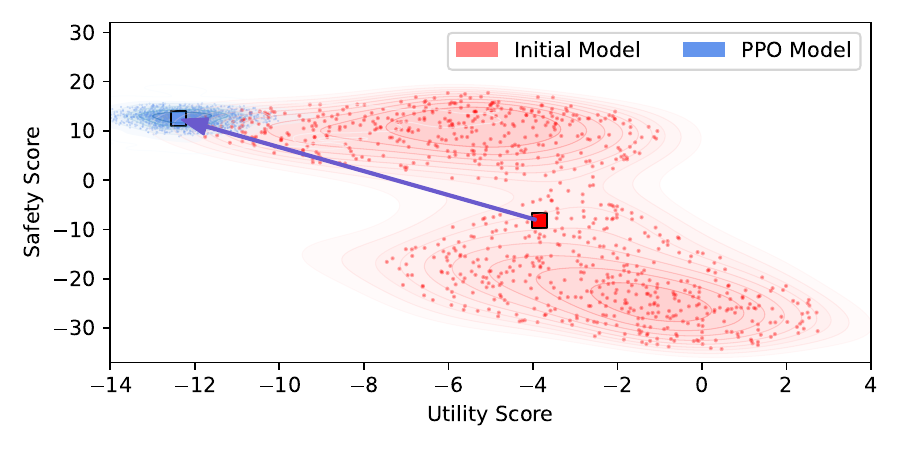}
        \caption{RLHF under Seq2scalar RM.}
    \end{subfigure}
    \begin{subfigure}[b]{0.48\textwidth}
        \centering
        \includegraphics[width=\textwidth]{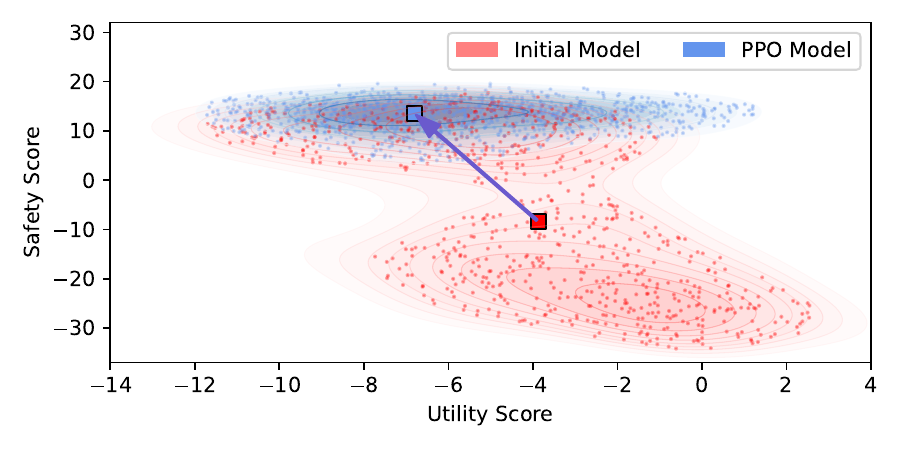}
        \caption{RLHF under Seq2seq RM.}
    \end{subfigure}
    \caption{Utility and safety scores distribution shifts of safety alignment. RLHF based on seq2scalar RM (upper) causes LLMs to misgeneralize the refusal-to-response paradigm. Our seq2seq RM (lower) mitigates unexpected generalization, improving safety while maintaining the utility scores distribution.
    }
    \label{fig:exp:dist}
\end{figure}

The seq2seq RM demonstrates a superior capability to represent preferences, effectively mitigating the refusal-to-response paradigm. PPO-S achieved a higher safety score compared to PPO while maintaining a slightly higher mean and variance in the utility score. This suggests that the seq2seq RM provides more accurate rewards in scalar space than the seq2scalar RM. Notably, PPO-T achieved the highest golden safety score without a significant decrease in the utility score. This indicates that the token-level fine-grained rewards of the seq2seq RM are crucial in mitigating unexpected generalization. 

A more intuitive visualization of utility score \emph{Vs.} proxy safety score has been presented in Figure \ref{fig:exp:dist}. Specifically, when the alignment objective is to improve the safety scores, \textit{i.e.}, from bottom to top in the diagram, the RLHF based on seq2scalar RM unexpectedly generalizes to refusal-to-response paradigm, \textit{i.e.}, from right to left, resulting in a significant drop and distribution collapse in the utility scores. In contrast, the seq2seq RM can improve the safety scores while maintaining the utility score distribution.

\subsection{Improvement of Alignment Perfromance}
\label{exp: efficiency}

In this section, we will demonstrate that the accuracy and fine-grained advantages of seq2seq RM can improve alignment efficiency. We empirically verify this improvement at both the token-level and the sentence-level optimization.

\noindent \paragraph{Sentence-Level Accuracy} Seq2seq RM can accurately distinguish between good and bad responses. We sum the token-level rewards provided by seq2seq RM up to scalars and conduct an empirical analysis on RFT and PPO. RFT prompts LLMs to generate multiple responses (set to 8 in our experiment), then selects the highest-scoring response for further SFT. Both methods' alignment performance relies on whether the RM can provide accurate sentence-level feedback. As shown in Table \ref{tab: efficiency}, both RFT and PPO using seq2seq RM achieved the best performance across different model sizes and tasks. This indicates that using sequence MLE in the language space instead of binary MLE in the scalar space improves the accuracy of reward modeling.

\noindent \paragraph{Token-Level Granularity} To verify the token-level accuracy of the seq2seq RM, we compared PPO-T with several popular alignment algorithms. In addition to PPO, we selected DPO and RFT as baseline methods. Our results show that PPO-T achieved SOTA performance, demonstrating that the seq2seq RM effectively provides token-level feedback. We also compared PPO-S with PPO-T, as they only differ in feedback granularity. The significant outperformance of PPO-T over PPO-S further indicates that the seq2seq RM excels in accurate credit assignment at the token level.

\noindent \paragraph{OOD Robustness} 

\begin{table}[htp]
  \centering
  \resizebox{\columnwidth}{!}{
\begin{tabular}{@{}lccccccc@{}}
\toprule
\multicolumn{1}{c}{} & \multicolumn{1}{c}{Gemma-2B} & \multicolumn{1}{c}{Llama2-7B} \\ \midrule
PPO \emph{Vs.} SFT        & 56.7\% $\pm$ 22.8\%  & 61.6\% $\pm$ 28.4\%  \\
PPO-T \emph{Vs.} SFT      & 75.2\% $\pm$ 18.7\%  & 74.2\% $\pm$ 19.1\%  \\
PPO-T \emph{Vs.} PPO      & 71.9\% $\pm$ 20.2\%  & 68.3\% $\pm$ 20.5\%  \\
\bottomrule
\end{tabular}
  }
  \caption{Comparison of methods for Gemma-2B and Llama2-7B on OOD Xsum \cite{narayan2018don} prompts, with RMs trained on TL;DR datasets.}
  \label{tab:xsum}
\end{table}

The seq2seq RM demonstrates robust scoring accuracy for out-of-distribution (OOD) responses. We tested the seq2seq RM and seq2scalar RM based on the TL;DR datasets but used the Xsum prompt in the RLHF process. Since the prompts in Xsum are not in the same distribution as those in TL;DR, this will test whether the RM is robust to OOD inputs. PPO-T can significantly improve the performance of SFT models, whereas PPO improvements are not obvious. Further, PPO-T also outperforms PPO, which is based on the seq2scalar RM.

\subsection{Ablation Study and Disscussion}
\label{exp: ablation}

\noindent \paragraph{Necessity of Positive and Negative Scores} The positive and negative scores are both essential for the seq2seq RM. We conducted ablation experiments on PKU-SafeRLHF under conditions where only negative scores or only positive scores were included. Our experiments indicate that when only negative scores are used, large language models (LLMs) tend to converge to a paradigm where no characters are generated. This occurred 68.5\% of the time with Gemma-2B and 86\% of the time with Llama2-7B. We hypothesize that generating nothing is easier for LLMs than generating proper tokens, as both outcomes avoid negative scores. The inclusion of positive scores mitigates this issue by incentivizing the generation of tokens consistent with the seq2seq RM.

\begin{figure}[t]
        \centering
        \includegraphics[width=\columnwidth]{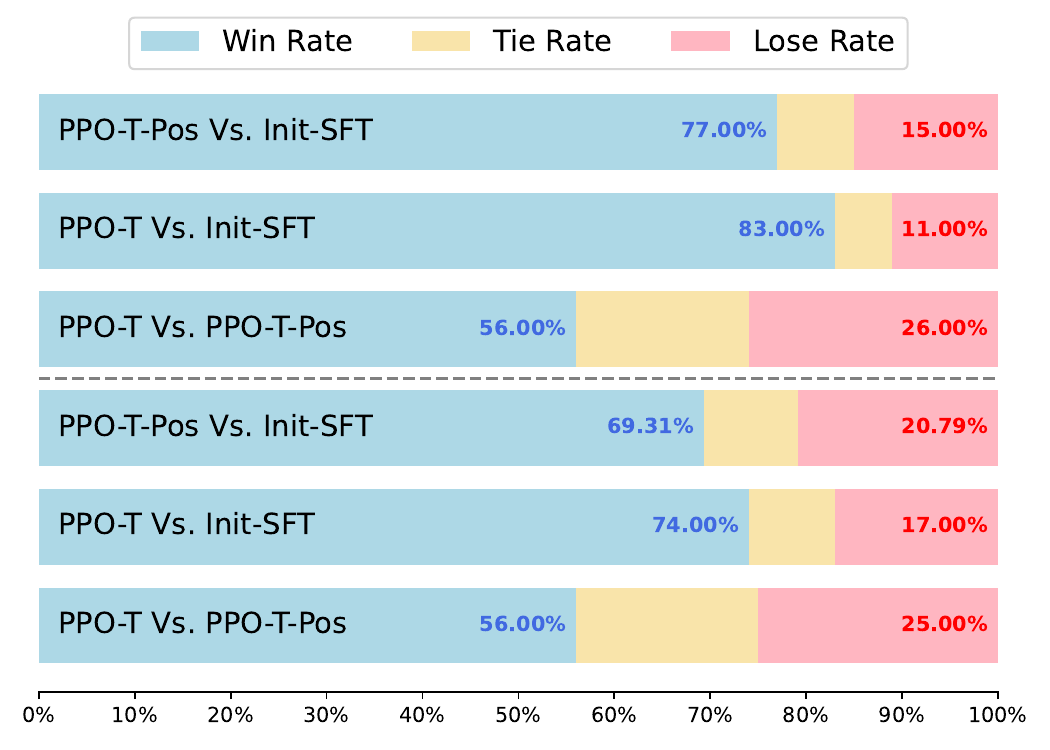}
    \caption{Comparison of PPO-T, PPO-T-Pos and Init-SFT. Bars above the dashed line are based on Gemma-2B while the others are based on Llama2-7B.}
    \label{fig: neg}
\end{figure}

Negative scores can enhance the alignment performance of seq2seq RMs. Figure \ref{fig: neg} compares the performance of three models: PPO-T, PPO-T-Pos (which considers only positive scores), and Init-SFT. The results demonstrate that PPO-T, which includes negative scores in its evaluation, significantly improves alignment performance compared to Init-SFT. This improvement suggests that incorporating negative scores allows for better utilization of information from the preference dataset, leading to more effective RL fine-tuning.

\noindent \paragraph{Correction Abilities of Seq2seq RM}

\begin{figure}[t]
        \centering
        \includegraphics[width=\columnwidth]{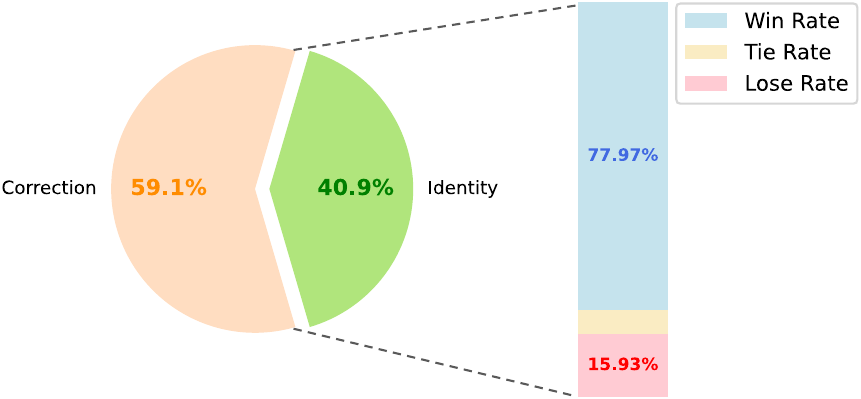}
    \caption{Quantified analysis of seq2seq RM's correction mapping and identity mapping. The pie chart shows the proportion of responses corrected and retained by seq2seq RM, while the bar chart shows the comparison results between the corrected responses and the original responses.}
    \label{fig: rm}
\end{figure}

\begin{table}[ht]
  \centering
  \resizebox{\columnwidth}{!}{
    \begin{tabular}{@{}lccc@{}}
    \toprule
    Models & PKU-SafeRLHF & Dialogsum & TL;DR \\ \midrule
    Gemma-2B & 76.7\% $\pm$ 17.9\% & 75.2\% $\pm$ 18.6\% & 73.2\% $\pm$ 12.2\% \\
    Llama2-7B & 82.2\% $\pm$ 14.6\% & 78.6\% $\pm$ 16.8\% & 79.3\% $\pm$ 15.1\%   \\ \bottomrule
    \end{tabular}
  }
  \caption{Win-lose ratio of PPO-T over Init-SFT responses corrected by seq2seq RM.}
  \label{tab: rm}
\end{table}

To verify that seq2seq RM has the ability to perform both correction mapping and identity mapping on responses, we concatenate the prompt with the response generated by Init-SFT and then input it into seq2seq RM. By comparing the output with Init-SFT, the results in Figure \ref{fig: rm} indicate that seq2seq RM is capable of mapping Init-SFT's responses to be more aligned with human preferences. 

However, relying solely on the combination of the Init-SFT model and seq2seq RM cannot surpass RLHF. As shown in Table \ref{tab: rm}, the performance of PPO-T based on seq2seq RM training is better than the direct correction of seq2seq RM. This indicates that the feedback from seq2seq RM enables LLMs to generalize to better alignment performance than simple correction mapping.

\section{Related Works}

\noindent \paragraph{Reward Modeling} Reward modeling is a crucial component of RLHF. Given that RM serves as imperfect proxies for human intentions, LLMs can exploit their vulnerabilities, exhibiting unexpected generalization to achieve high rewards during RL finetuning \citep{pitis2023failure, gao2023scaling}. It means RM fails to provide feedback that accurately aligns with human intentions: LLMs exploring unexpected generalization to obtain high rewards, which negatively impact their capabilities and fail to achieve alignment objectives \citep{tien2022causal, knox2023models, lambert2023alignment}. \citet{moskovitz2023confronting} proposed the use of Lagrange multipliers to balance rewards representing various preferences, aiming to create a more nuanced reward structure. Similarly, \citet{coste2023reward} explored ensembling multiple RMs to provide more accurate feedback by leveraging diverse perspectives. \citet{rame2024warm} involved averaging RM weights under different random settings to enhance robustness against variability. Additionally, \citet{zhu2024iterative} introduced iterative data refinement techniques to minimize noise within the reward signals. While these methods have demonstrated varying degrees of effectiveness, they all introduce additional models, training requirements, and hyperparameters, which may impact their robustness and scalability.

\noindent \paragraph{Reinforcement Learning from Human Feedback}

Recent research on token level RLHF presents an impossible triangle of \textit{granularity}, \textit{accuracy}, and \textit{annotation cost}. \citet{chan2024dense} redistributed the seq2scalar RM's output to each token based on attention values to achieve fine-grained feedback, but this still relies on the Bradley-Terry model's accuracy over whole responses in learning preferences. \citet{chen2024improving}, from the perspective of minimum editing constraint, used a generative model to provide token-level reward feedback, but this requires an additional teacher model for correction annotations. \citet{rafailov2024r} and \citet{zeng2024token} demonstrated the feasibility of achieving token-level alignment from a variant of RLHF, direct preference optimization (DPO) \citep{rafailov2024direct}. However, DPO's offline and off-policy setting reduces its alignment efficiency and cannot be optimized when the preference dataset variability is insignificant \citep{tajwar2024preference}. \citet{xu2024dpo} further showed through theoretical and experimental analysis that DPO is a suboptimal alternative to RLHF methods in various scenarios. This also raises concerns about using DPO as the reward model for RLHF by \citet{zhong2024dpo}.

\section{Conclusion}

In this work, we propose a novel sequence-to-sequence (seq2seq) reward modeling method. Its key insight is that learning from language feedback rather than scalar feedback improves RLHF without additional annotations. We replace the reward modeling target from binary MLE for classification tasks to sequence MLE for text generation tasks. Our experiments demonstrate that seq2seq RM reduces the refusal-to-response paradigm in single-turn safety dialogues and the long-response bias in text summarization tasks, improving RLHF performance across 2B and 7B LLMs, achieving an average win rate of 76.9\%. We further show that seq2seq RM can still improve the performance of RLHF under the out-of-distribution prompts.

\section{Acknowledgments}

This work is sponsored by National Natural Science Foundation of China (62376013, 623B2003)and Beijing Municipal Science \& Technology Commission (Project ID: Z231100007423015)

\bibliography{aaai25}

\end{document}